\documentclass[conference]{IEEEtran}
\IEEEoverridecommandlockouts
\usepackage{cite}
\usepackage{amsmath,amssymb,amsfonts}
\usepackage{algorithmic}
\usepackage{graphicx}
\usepackage{textcomp}
\usepackage{xcolor}
\usepackage{cuted}
\usepackage{subcaption}
\usepackage{booktabs}       
\usepackage[compatibility=false]{caption}
\usepackage{tabularx}
\usepackage{hyperref} 
\usepackage{fontawesome5}
\def\BibTeX{{\rm B\kern-.05em{\sc i\kern-.025em b}\kern-.08em
    T\kern-.1667em\lower.7ex\hbox{E}\kern-.125emX}}
\begin{document}

\title{Physics-informed Ground Reaction Dynamics \\ from Human Motion Capture}

\makeatletter
\newcommand{\linebreakand}{%
  \end{@IEEEauthorhalign}
  \hfill\mbox{}\par
  \mbox{}\hfill\begin{@IEEEauthorhalign}
}
\makeatother

\author{
\IEEEauthorblockN{Cuong Le}
\IEEEauthorblockA{\textit{Dept. of Electrical Engineering} \\
\textit{Linköping University}\\
Linköping, Sweden \\
cuong.le@liu.se}
\and
\IEEEauthorblockN{Huy-Phuong Le}
\IEEEauthorblockA{\textit{Fac. of Eletrical and Electronics} \\
\textit{Engineering, HCMUTE University}\\
Ho Chi Minh City, Vietnam \\
phuonglehuy172k@gmail.com}
\and
\IEEEauthorblockN{Duc Le}
\IEEEauthorblockA{\textit{Fac. of Electrical and Electronics} \\
\textit{Engineering, HCMUTE University}\\
Ho Chi Minh City, Vietnam \\
hienduc.spk@gmail.com}
\linebreakand
\IEEEauthorblockN{Minh-Thien Duong}
\IEEEauthorblockA{\textit{Dept. of Automatic Control} \\
\textit{HCMUTE University}\\
Ho Chi Minh City, Vietnam \\
minhthien@hcmute.edu.vn}
\and
\IEEEauthorblockN{Van-Binh Nguyen}
\IEEEauthorblockA{\textit{Inst. of Engineering-Technology} \\
\textit{Thu Dau Mot University}\\
Thu Dau Mot City, Vietnam \\
binhnv@tdmu.edu.vn}
\and
\IEEEauthorblockN{My-Ha Le}
\IEEEauthorblockA{\textit{Fac. of Electrical and Electronics} \\
\textit{Engineering, HCMUTE University}\\
Ho Chi Minh City, Vietnam \\
halm@hcmute.edu.vn}
}

\maketitle

\begin{strip}
    \vspace{-40pt}
    \centering
    \captionsetup{type=figure}
    \includegraphics[width=\textwidth,height=4.5cm]{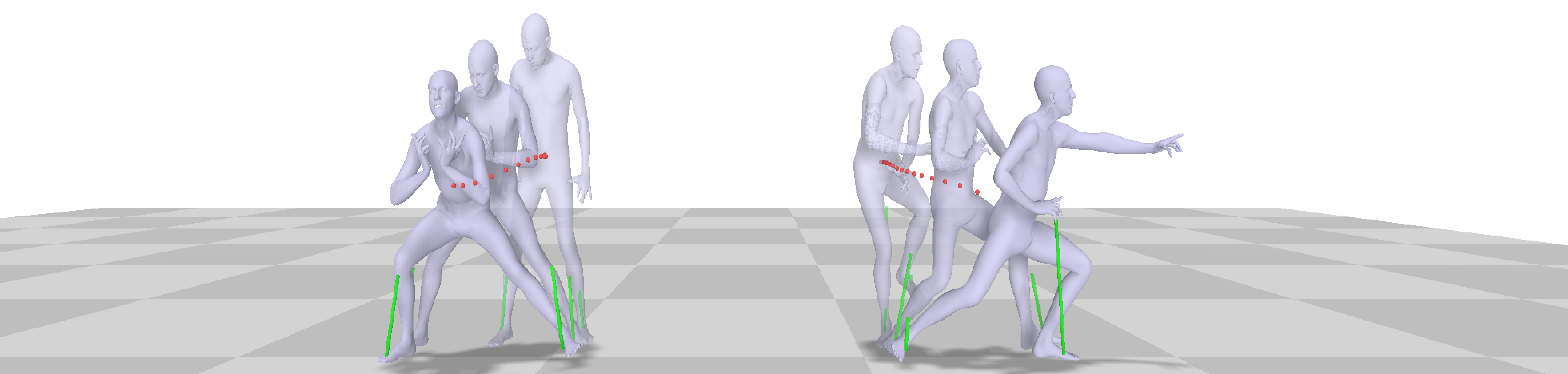}
    \captionof{figure}{The overview of the proposed physics-informed ground reaction dynamics estimation from human motion capture. The reactive dynamics (green) is calculated directly from the global translation (red) of the body according to physics laws. This formulation brings a more robust and reliable form of supervision, in addition to force plate data, to any machine learning model that predicts ground reaction dynamics.
    \label{fig:teaser}}
\end{strip}

\begin{abstract}
    Body dynamics are crucial information for the analysis of human motions in important research fields, ranging from biomechanics, sports science to computer vision and graphics.
    Modern approaches collect the body dynamics, external reactive force specifically, via force plates, synchronizing with human motion capture data, and learn to estimate the dynamics from a black-box deep learning model.
    Being specialized devices, force plates can only be installed in laboratory setups, imposing a significant limitation on the learning of human dynamics.
    To this end, we propose a novel method for estimating human ground reaction dynamics directly from the more reliable motion capture data with physics laws and computational simulation as constrains.
    We introduce a highly accurate and robust method for computing ground reaction forces from motion capture data using Euler's integration scheme and PD algorithm.
    The physics-based reactive forces are used to inform the learning model about the physics-informed motion dynamics thus improving the estimation accuracy.
    The proposed approach was tested on the GroundLink dataset, outperforming the baseline model on: 1) the ground reaction force estimation accuracy compared to the force plates measurement; and 2) our simulated root trajectory precision.
    The implementation code is available at \href{https://github.com/cuongle1206/Phys-GRD}{\faIcon{github}}.
\end{abstract}


\section{Introduction}
\label{sec:introduction}

Human dynamics are the relevant physical factors responsible for the human body movements such as foot-ground contact position, body velocity and acceleration, and especially ground plane reaction forces.
The retrieval of these information is a crucial part for fields that require intensive understanding of human body movements, i.e. sport analysis \cite{arboix_sport24,kadlec_sport24}, biomechanics \cite{derlatka_gait23,park_gait22}, or computer graphics \cite{starke_anim19,han_groundlink23}.

Despite being the vital part of many related studies, capturing human ground reactive dynamics remains a challenging problem due to the unobservable nature of the force variables.
Traditional approach often make use of external devices such as force plates for the collection of ground reaction forces.
Force plates are special sensors that could measure the direction and magnitude the force and moment applying to the surface of the devices \cite{beckham_plate14}.
Despite the valuable measurements, force plates require a very accurate laboratory setup and actor's performance to achieve usable measurements.
Furthermore, the data collected from force plates is prone to discontinuity, since the plates are often installed in a restricted region within the recording space, and not available in other areas.
Motion capture data, on the other hand, is available in a much wider range and does not require the performer to follow special instructions, leading to the capture of more natural movements.

We propose a novel approach for estimating the ground reactive dynamics directly from motion capture data.
The total ground reaction force, together with the gravitational force, is responsible for the global translation of the human body in world space.
We utilize the Proportional-Derivative (PD) algorithm to estimate the physics-based reactive forces proportionally to the offset between two consecutive body root joint positions.
Newtonian physics law is then enforced on the estimated forces to obtain a simulated root trajectory, and the PD control gains are selected to minimize the error of the simulated results.
The physics-based reactive forces serve as an additional training objective for the deep learning model, increasing the physics plausibility of the model's predictions.

The approach is trained and evaluated against a baseline model on the GroundLink dataset \cite{han_groundlink23}, taking motion capture data as inputs to estimate ground reaction forces (GRF) and contact center of pressure (CoP).
Our main contributions are listed as the followings:
\begin{itemize}
    \item We propose a new approach for directly estimate accurate ground reaction forces from motion capture data, reducing the dependency on the limited force plates data,
    \item We demonstrate the evaluation metric for ground reaction force prediction based on motion capture data is more reliable than prone-to-discontinuation force plates data,
    \item We show that the deep learning model trained with the additional physics-informed objective is more accurate than the baseline model.
\end{itemize}

\section{Related works}
\label{sec:related_works}

\subsection{Human motion capture}
\label{subsec:related_mocap}

Capturing human motion is a long-standing research especially in the field of computer vision.
The task often involves the retrieval of human 2D joints \cite{alphapose,xu_vitpose22} in camera space, or 3D joints \cite{holmquist_diffpose24,peng_ktp24} in world space.
Despite the important role in tasks such as tracking or behavior analysis, 2D/3D human poses are less informative than physical dynamics for studies of human biomechanics or sport analysis \cite{uchida_bio20}. This leads to an emerging field of capturing the dynamics from human motion, including the estimation of contacts and external reactive forces from the surrounding environment.

\subsection{Human dynamics capture}
\label{subsec:related_dynamics}

Human physical dynamics are difficult to obtain because it is unobservable directly from the captured motions.
Prior biomechanics research often measure the explicit reaction forces and moments via specialized devices such as force plates \cite{camargo_lower21,zee_walking22}.
However, these methods are usually limited to locomotion (walking or running) and cannot generalize well to different type of motions. Some other approaches for collecting ground reactive dynamics is via pressure insoles \cite{jesse_eccv20,mourot_cgf22} or internal measurement units \cite{wang_wearable23}.
These approaches provide a more accurate measurement on reactive dynamics, especially contact states, but often required intrusive sensors, causing unnatural performance of the actors and limiting the usage case to only laboratory setting.
Recently, \cite{han_groundlink23} introduces the GroundLink dataset, which consists of a wide range of motion data along with synchronizing explicit ground reaction measurements from force plates, enabling more in-depth research on human dynamics from motion capture.
Explicit dynamics, despite their valuable information, are unreliable long-term measurements due to the limitation of the sensor laboratory setup.
We address this issue by incorporating a physics-based information from motion capture data into the prediction of ground reaction dynamics model.

Unlike force plates' measurements, motion capture is more robust and widely adaptable source of information.
This characteristic stems the emerging field of physics-based human motion capture that implicitly estimates the human dynamics directly from motion capture data \cite{shimada_physcap20,shimada_neurphys21,le_osdcap24, zell_eccv20}.
Modern vision-based motion capture approaches adapt physics-based information to refine their noisy human pose estimation, often resulting in physical dynamics as the by-products.
\cite{shimada_physcap20} estimates the human body dynamics via an trajectory optimization approach.
The method optimizes the dynamics variable such that they minimize the objective loss based on physics laws.
Trajectory optimization approaches are however cannot directly be applied to new data, requiring a full expensive optimization process.
\cite{shimada_neurphys21} addresses this problem by shifting the estimation of human dynamics to a data driven approach, by employing a learning model \emph{GRFNet} for ground reaction prediction.
The human character used to compute physics constraints in \cite{shimada_neurphys21} is however self-created based on body shape simple approximation, leading to incorrect dynamics predictions with respect to the observing human, as pointed out by \cite{le_osdcap24}.
We, on the other hand, utilize the corresponding real measurements from force plates of \cite{han_groundlink23} together with the physics-based calculation to achieve the accurate results.

\section{Method}
\label{sec:method}

\begin{figure}[t]
    \centering
    \includegraphics[width=0.98\linewidth]{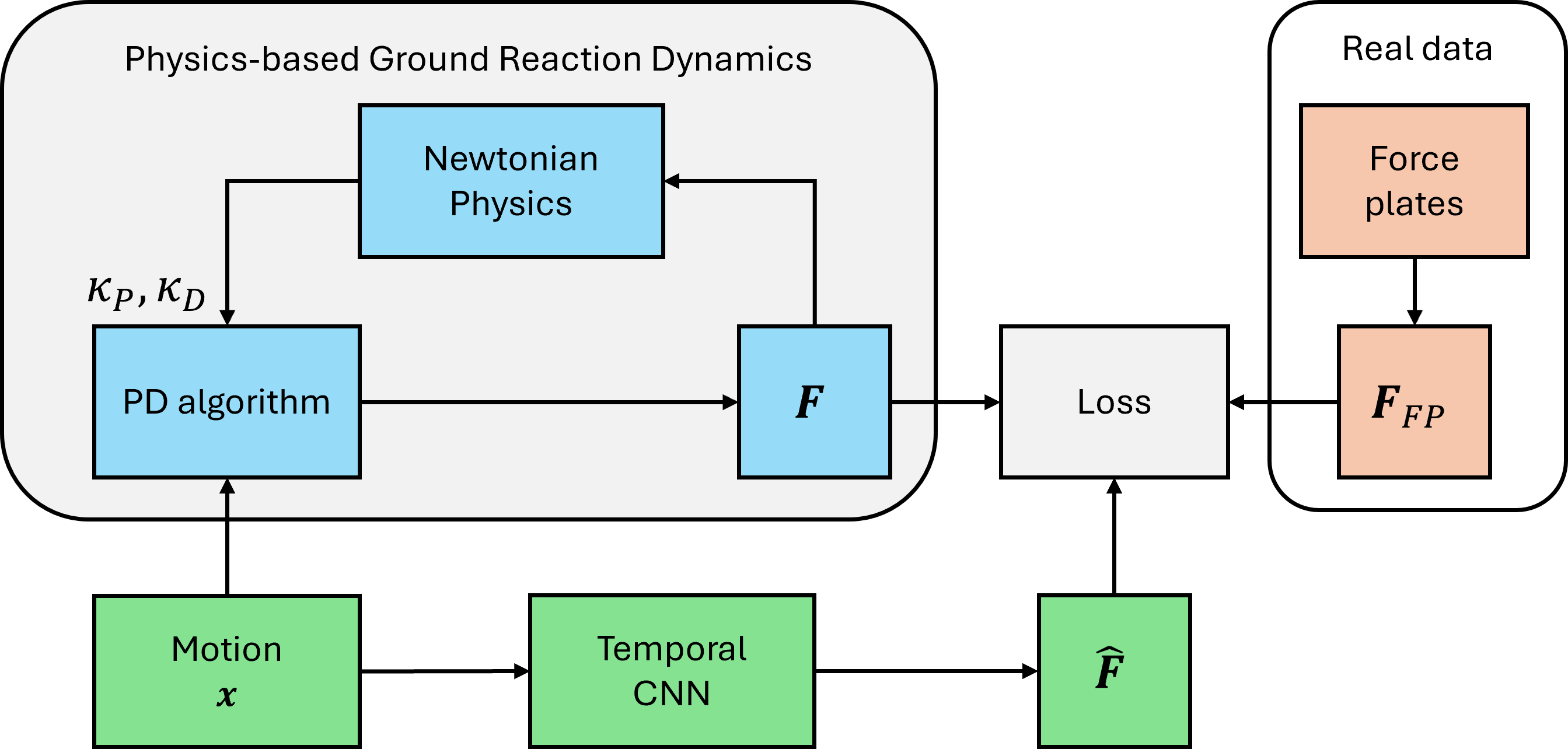}
    \caption{The pipeline of the proposed approach. The motion capture data is used as the input for the temporal convolution model to predict the ground reaction forces. Two sources of supervision are provided: 1) real data collected from force plates synchronizing with the motion capture data; and 2) the physics-based reaction forces computed directly from the input motion using PD algorithm. The control parameters of the PD algorithm is determined by minimizing the reconstruction error between the simulated trajectory and the input motion.}
    \label{fig:pipeline}
\end{figure}

Our aim is to reinforce the data-driven prediction of ground reaction forces with a physics-based pseudo ground truth computed directly from the motion capture data. An overview of the proposed approach can be found in Fig. \ref{fig:pipeline}.

\subsection{Physics-based trajectory simulation}
\label{subsec:method_traj}

To facilitate the physics-based calculation of ground reaction forces, we utilize the concept of residual forces that assumes all external forces acting on the human body during motions are responsible for the global translation of the whole body \cite{shimada_physcap20}. This means the ground reaction forces $\mathbf{F} \in \mathbb{R}^3$, together with the gravitational force $\mathbf{G} \in \mathbb{R}^3$, control the movement of the human body in world space, under the Newtonian physics law of motions. Given the root 3D position of the body root as $\mathbf{x} \in \mathbb{R}^3$, the Newtonian equation of motion is defined as:
\begin{equation}
    \mathbf{F} - \mathbf{G} = m \Ddot{\mathbf{x}},
    \label{eq:eom}
\end{equation}
where $m$ is the total mass of the human body and $\mathbf{\Ddot{x}}$ is the acceleration of the root body joint in world coordinate. The subtraction between $\mathbf{F}$ and $\mathbf{G}$ denotes the opposite applied direction of the two force vectors. While the gravitational force $\mathbf{G}$ is assumed to be $[0., 0., 9.81]$, the $\mathbf{F}$ can be approximated with PD algorithm given a trajectory, following \cite{shimada_neurphys21, le_osdcap24}:
\begin{equation}
    \mathbf{F}_t = \kappa_P (\mathbf{x}_{t+1}-\mathbf{x}_{t}) - \kappa_D \dot{\mathbf{x}}_{t},
    \label{eq:pd}
\end{equation}
where $\kappa_P$ and $\kappa_D$ are PD gains that control the magnitude of the applied reaction force $\mathbf{F}$ with respect to the offset between root joint position of two consecutive frames ($\mathbf{x}_{t}$ and $\mathbf{x}_{t+1}$), together with a dampening factor proportional to $\dot{\mathbf{x}_{t}}$.

To obtain the correct control parameters $\kappa_P$ and $\kappa_D$, a simulation process is required to create a simulated trajectory, given the estimated ground reaction force $\mathbf{F}$ from Eq. \ref{eq:pd}. The simulation is based on the Euler's numerical integration scheme and is given as:
\begin{equation}
    \begin{split}
        & \dot{\mathbf{x}}_{t+1} = \dot{\mathbf{x}}_{t} + \Ddot{\mathbf{x}}_{t} \Delta t, \\
        & \hat{\mathbf{x}}_{t+1} = \hat{\mathbf{x}}_{t} + \dot{\mathbf{x}}_{t+1} \Delta t,
    \end{split}
    \label{eq:euler}
\end{equation}
where $\Delta t$ is the simulation step size and $\hat{\mathbf{x}}$ is the simulated root trajectory given the calculated reaction force $\mathbf{F}$. The parameters $\kappa_P$ and $\kappa_D$ are chosen to minimize the mean squared error between the simulated trajectory $\hat{\mathbf{x}}$ and the original motion capture data $\mathbf{x}$. An example of a good PD control parameter search is shown in Fig. \ref{subfig:hopping}, where the vertical root trajectory matches the original data and the corresponding vertical reaction force is close to the measurement data from the force plates.

As discussed in Sec. \ref{sec:related_works}, force plates data often contain missing measurements as can be seen in Fig. \ref{subfig:walk}, especially in the beginning of the trials, potentially leading to a false analysis afterward. Having a physics-based calculation help reinforce these cases, providing the system with more robust and consistent measurements of human dynamics.

\subsection{Physics-informed GRFs estimation}
\label{subsec:method_grf}

Similar to the baseline approach \cite{han_groundlink23}, we also treat the task of predicting the ground reaction dynamics as a black-box process. As illustrated in Fig. \ref{fig:pipeline}, the motion capture data is used as inputs to the deep learning model, which consists of four layers of temporal convolutions with activation function ELU, followed by three fully-connected layers to estimate the ground reaction force vector $\hat{\mathbf{F}}$ in world coordinate. By integrating physics knowledge into the model, the learning objective now contains two loss terms: 1) the mean squared error between the prediction and the measurement from force plates' data; and 2) the error between the prediction and the physics-based estimation of total ground reaction force from Newtonian physics. The objective function $\mathcal{L}$ takes the form:
\begin{equation}
    \mathcal{L} = \frac{1}{T} \sum^{T}_{t=1} \lambda_1 \lVert \mathbf{F}_{FP} - \hat{\mathbf{F}} \rVert ^{2} + \lambda_2 \lVert \mathbf{F} - \sum^{2}_{k=1} \hat{\mathbf{F}} \rVert ^{2},
    \label{eq:loss}
\end{equation}
where $\mathbf{F}_{FP}$ is the ground reaction data collected from force plates, $T$ is the total of frames for one sample of human motion, $\lambda_1$ and $\lambda_2$ are the learning weights for the respective loss terms. Since the physics-based $\mathbf{F}$ is the total ground reactivate force projected to the root joint, $k$ denotes the number of contacts ($2$ for two legs in our case).

\begin{figure*}[!t]
    \centering
    \begin{subfigure}[h]{0.45\textwidth}
        \centering
        \includegraphics[width=\linewidth]{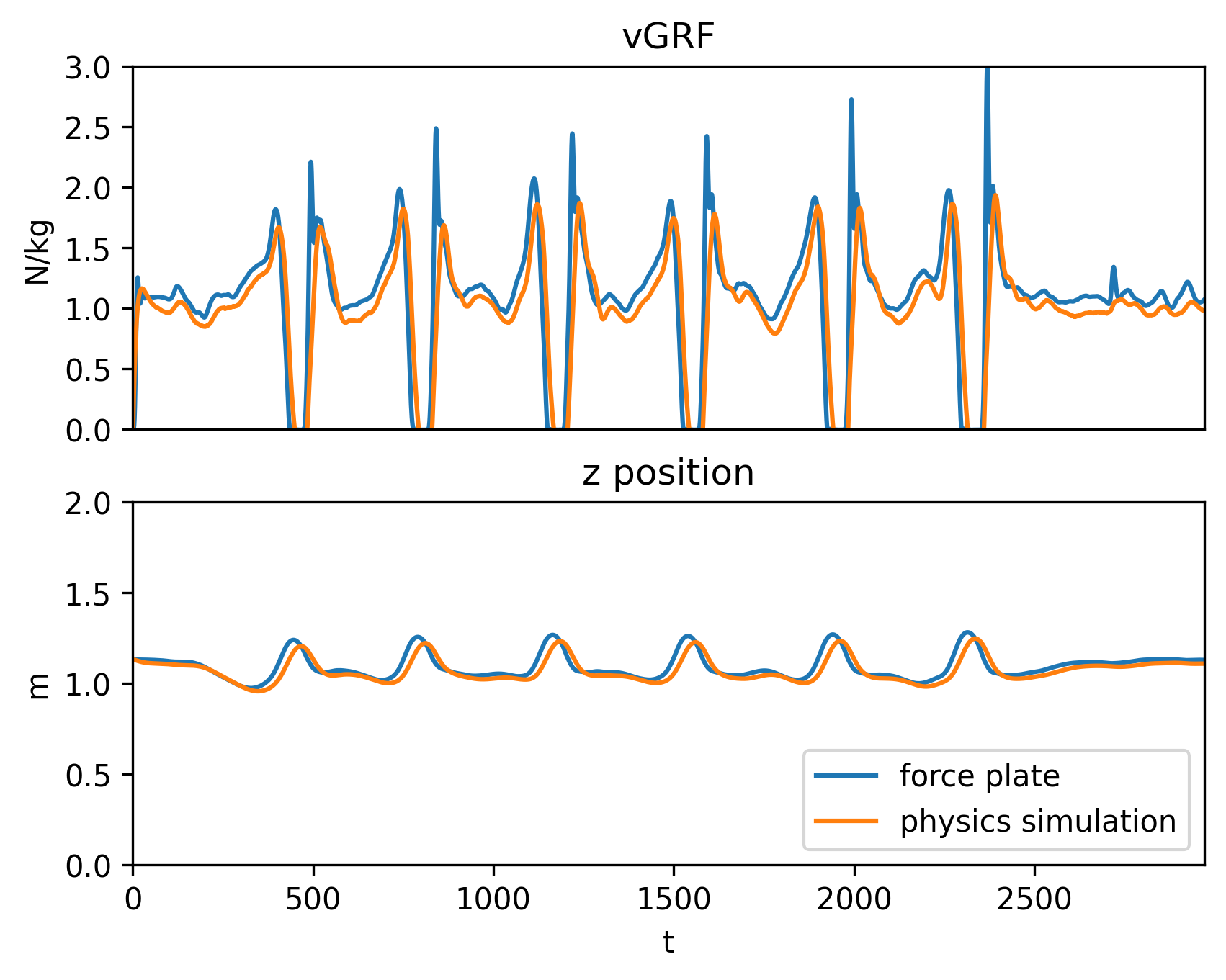}
        \caption{Stationary Hopping}
        \label{subfig:hopping}
    \end{subfigure}
    ~
    \begin{subfigure}[h]{0.45\textwidth}
        \centering
        \includegraphics[width=\linewidth]{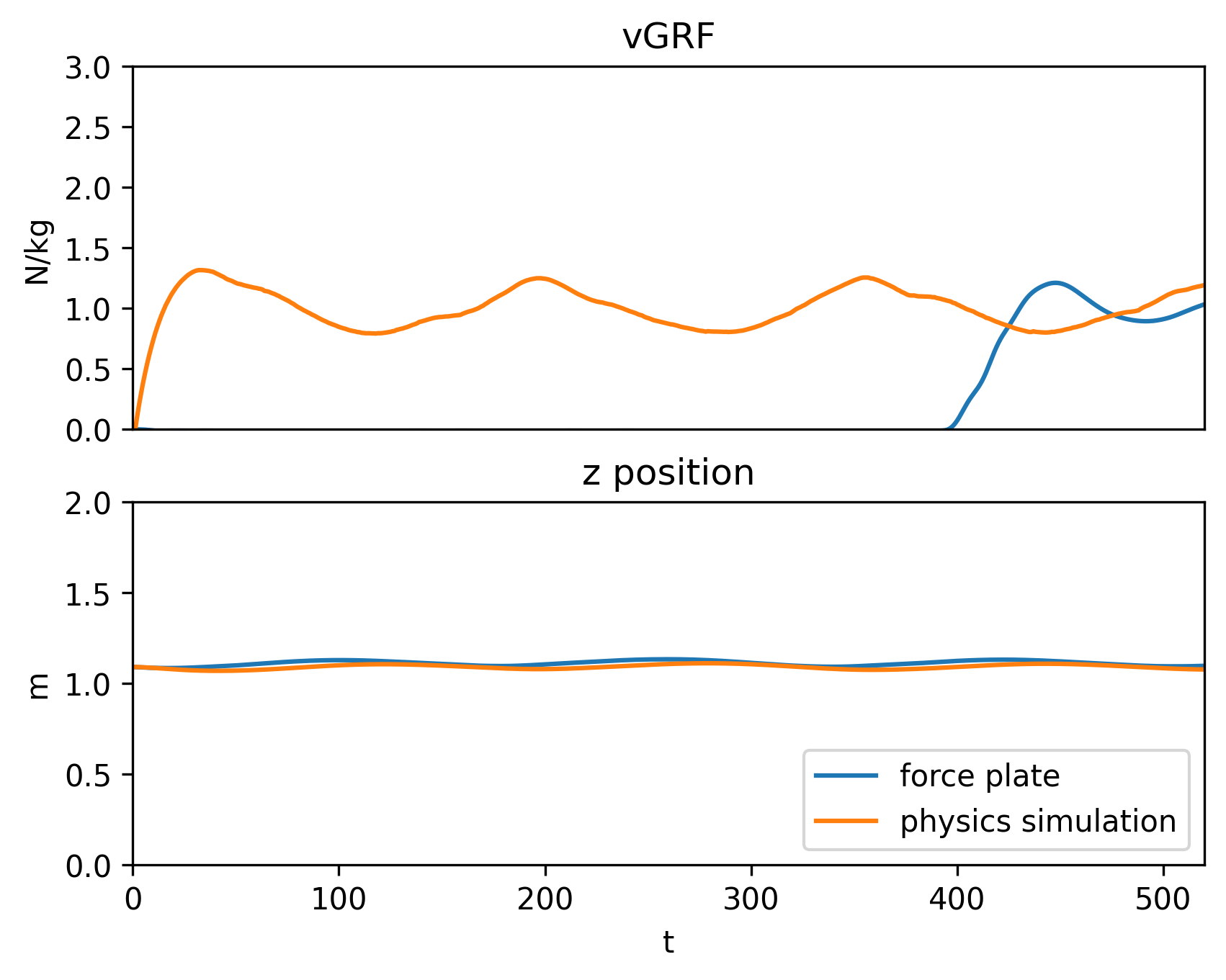}
        \caption{Walking}
        \label{subfig:walk}
    \end{subfigure}
    \caption{An example of physics-based trajectory simulation (orange) vs. measurement data from force plates (blue) on the stationary hopping and walking motions. With a good selection of control parameters in Eq. \ref{eq:pd}, the physics-based ground reaction dynamics can produce an accurate simulated root trajectory with respect to the original motion capture data. Force plates often encounter missing data as in \ref{subfig:walk}, causing incorrect measurements. Having no limitation on the sensor configuration, the physics-based reactive forces provide a more robust observation of human dynamics compared to force plates.}
    \label{fig:phys_sim}
\end{figure*}

\section{Experiments}
\label{sec:experiments}

\subsection{Dataset}
\label{subsec:dataset}

We evaluate our method on the GroundLink dataset \cite{han_groundlink23}.
The dataset contains diverse human motion capture data from seven different actors.
Every motion trial has full body kinematics with corresponding GRF and CoP with contact annotations.
The dataset consists of $19$ different subtle movements, ranging from locomotion to complex movements such as taichi, lambada dance, and jumping jack.
The capture motions are provided in SMPL pose parameters $\theta$ and $\beta$ obtained by Mosh++ \cite{amass19} on the markers position.
We only use the pose $\theta$ as the input for the learning model.
The first degree-of-freedom of $\theta$ is the global translation $\mathbf{x}$ in Eq. \ref{eq:pd}, and we use $\mathbf{x}$ for the calculation of the physics-based ground reaction dynamics.

\subsection{Implementation details}
\label{subsec:implemetation}

To keep a fair comparison, we adopt the \emph{GroundLinkNet} from \cite{han_groundlink23}, which consists of four 1D temporal convolution layers with kernel size of $7$.
The embedded variable is followed by three fully connected layers that adapt to variable-length sequences.
The activation function is exponential linear unit (ELU) and is applied after each convolution and fully connected layer.
The model is optimized with respect to the loss proposed in Sec. \ref{subsec:method_grf}.
All experiments are trained with a total of $11$ epochs with batch size of $64$ and learning rate of $3e^{-5}$, with the chosen random seed of $42$.
Cross-validation is applied to evaluate the results, using one participant as testing data and the rest for training.

\begin{table}[t]
    \centering
    \caption{Quantitative results of the proposed approach on vGRF in comparison with related works. The report errors are the MSE between the model's prediction and the force plates data. The values are shown for the two foot contacts ($\text{left} \, \vert \, \text{right}$). $^{\dagger}$The results are taken from \cite{han_groundlink23}.}
    \begin{tabularx}{\linewidth}{@{\extracolsep{\fill}}l|ccc}
        \toprule
        Motion & UnderPressure$^{\dagger}$ & GroundLink & Ours \\
        \midrule
        Chair               & $0.94 \, \vert \, 0.19$ & $0.03 \, \vert \, 0.02$ & $0.01 \, \vert \, 0.01$\\
        Tree (arm up)       & $0.54 \, \vert \, 0.11$ & $0.19 \, \vert \, 0.01$ & $0.20 \, \vert \, 0.01$\\
        Tree (arm down)     & $0.78 \, \vert \, 0.10$ & $0.17 \, \vert \, 0.01$ & $0.06 \, \vert \, 0.01$\\
        Warrior I           & $0.15 \, \vert \, 0.16$ & $0.01 \, \vert \, 0.04$ & $0.01 \, \vert \, 0.02$\\
        Warrior II          & $0.09 \, \vert \, 0.09$ & $0.01 \, \vert \, 0.06$ & $0.02 \, \vert \, 0.02$\\
        Dog                 & $0.69 \, \vert \, 0.13$ & $0.06 \, \vert \, 0.04$ & $0.09 \, \vert \, 0.08$\\
        Side Stretch        & $0.31 \, \vert \, 0.46$ & $0.07 \, \vert \, 0.06$ & $0.03 \, \vert \, 0.03$\\
        Lambada Dance       & $0.27 \, \vert \, 0.40$ & $0.05 \, \vert \, 0.14$ & $0.05 \, \vert \, 0.09$\\
        Squat               & $0.16 \, \vert \, 0.14$ & $0.04 \, \vert \, 0.02$ & $0.03 \, \vert \, 0.02$\\
        Jumping Jack        & $0.60 \, \vert \, 0.50$ & $0.06 \, \vert \, 0.16$ & $0.18 \, \vert \, 0.23$\\
        Stationary Hopping  & $0.80 \, \vert \, 0.27$ & $0.29 \, \vert \, 0.03$ & $0.22 \, \vert \, 0.05$\\
        Taichi              & $0.19 \, \vert \, 0.18$ & $0.03 \, \vert \, 0.08$ & $0.02 \, \vert \, 0.03$\\
        Soccerkick          & $0.49 \, \vert \, 0.41$ & $0.34 \, \vert \, 0.10$ & $0.18 \, \vert \, 0.09$\\
        Ballet Jump         & -                       & $0.16 \, \vert \, 0.07$ & $0.11 \, \vert \, 0.05$\\
        Walk                & -                       & $0.18 \, \vert \, 0.12$ & $0.18 \, \vert \, 0.17$\\
        Idling              & -                       & $0.04 \, \vert \, 0.04$ & $0.01 \, \vert \, 0.02$\\
        \midrule
        Average             & $0.44 \, \vert \, 0.23$ & $0.11 \, \vert \, 0.06$ & $\bold{0.09} \, \vert \, 0.06$\\
        \bottomrule
    \end{tabularx}
    \label{tab:vGRF_results}
\end{table}

\begin{table}[t]
    \centering
    \caption{Quantitative results of the proposed approach on vRPE in comparison with the baseline GroundLink. The report errors are the MSE between the root trajectory reconstruction from the model's ground reaction prediction and the original motion capture data. The results are computed in meters and scaled up by a factor of $10^3$.}
    \begin{tabularx}{\linewidth}{@{\extracolsep{\fill}}l|rr}
        \toprule
        Motion & GroundLink & Ours \\
        \midrule
        Chair               & $16.8  \pm 2.40$ & $4.45  \pm 1.11$ \\
        Tree (arm up)       & $34.75 \pm 3.41$ & $44.54 \pm 2.76$ \\
        Tree (arm down)     & $29.72 \pm 3.32$ & $9.17  \pm 3.47$ \\
        Warrior I           & $13.28 \pm 0.10$ & $2.65  \pm 0.31$ \\
        Warrior II          & $16.92 \pm 1.13$ & $3.61  \pm 1.04$ \\
        Dog                 & $57.82 \pm 4.93$ & $7.76  \pm 1.91$ \\
        Side Stretch        & $40.22 \pm 8.57$ & $4.18  \pm 0.23$ \\
        Lambada Dance       & $25.42 \pm 4.52$ & $9.90  \pm 2.32$ \\
        Squat               & $30.27 \pm 1.34$ & $14.22 \pm 0.83$ \\
        Jumping Jack        & $45.37 \pm 4.73$ & $40.04 \pm 5.77$ \\
        Stationary Hopping  & $45.62 \pm 7.51$ & $27.99 \pm 4.80$ \\
        Taichi              & $22.97 \pm 0.64$ & $7.19  \pm 2.13$ \\
        Soccerkick          & $84.97 \pm 9.66$ & $25.21 \pm 7.13$ \\
        Ballet Jump         & $47.42 \pm 1.15$ & $19.41 \pm 2.63$ \\
        Walk                & $75.59 \pm 42.39$ & $7.9  \pm 4.67$ \\
        Idling              & $19.29 \pm 1.37$ & $2.85  \pm 1.92$ \\
        \midrule
        Average             & $38.43 \pm 22.56$ & $\bold{14.69 \pm 13.37}$ \\
        \bottomrule
    \end{tabularx}
    \label{tab:vRPE_results}
\end{table}

\begin{figure*}[t]
    \centering
    \begin{subfigure}[h]{0.23\textwidth}
        \centering
        \includegraphics[width=\linewidth]{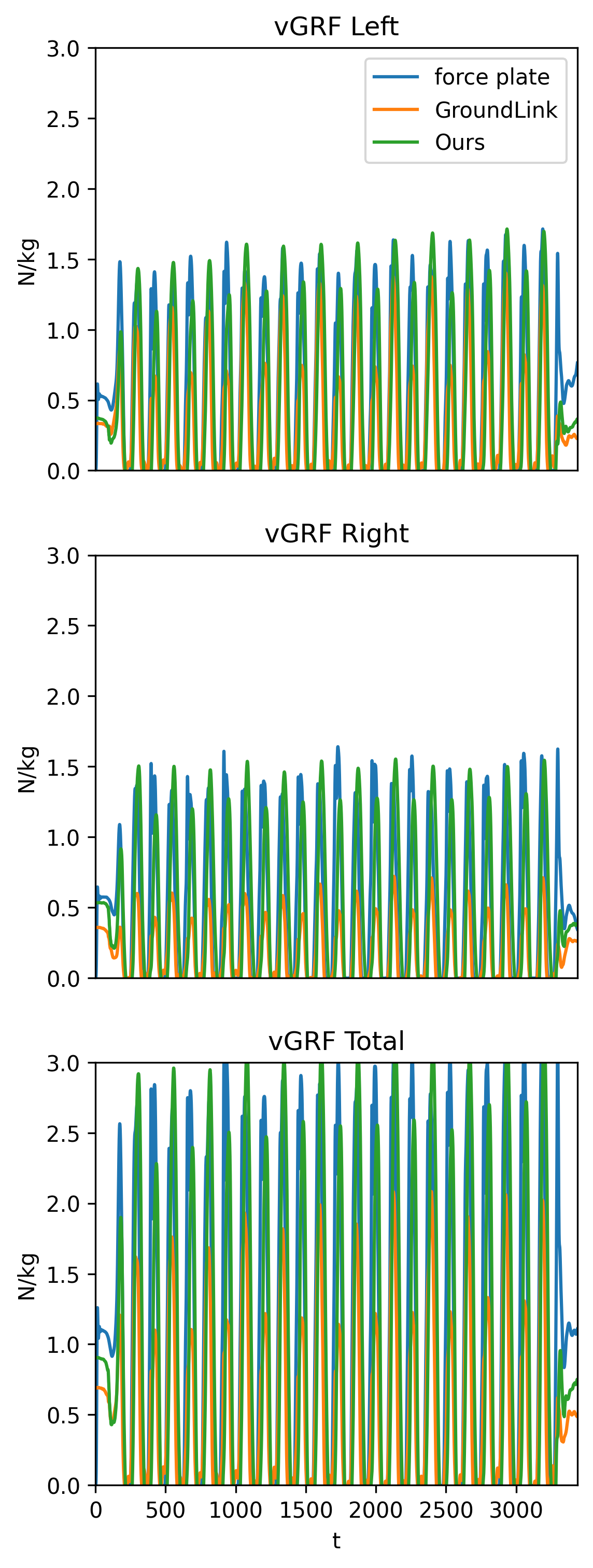}
        \caption{Jumping Jack}
        \label{subfig:jack}
    \end{subfigure}
    ~
    \begin{subfigure}[h]{0.23\textwidth}
        \centering
        \includegraphics[width=\linewidth]{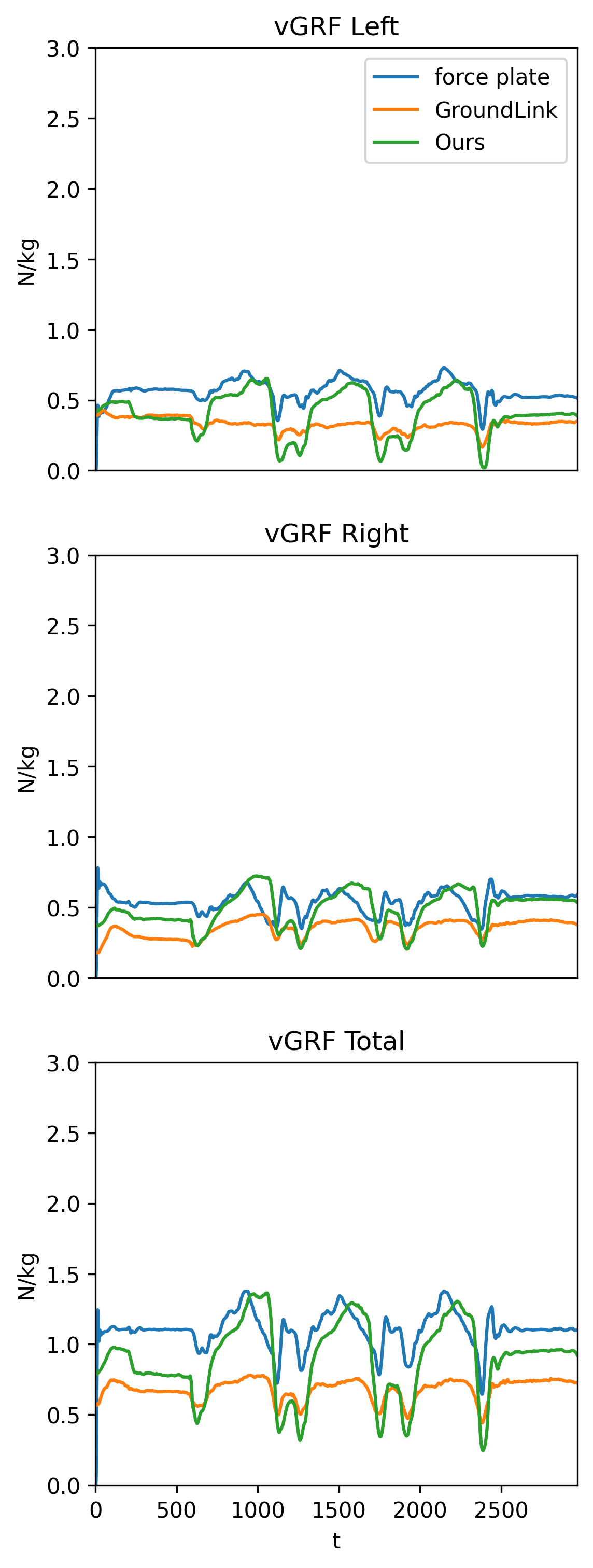}
        \caption{Squatting}
        \label{subfig:squat}
    \end{subfigure}
    ~
    \begin{subfigure}[h]{0.23\textwidth}
        \centering
        \includegraphics[width=\linewidth]{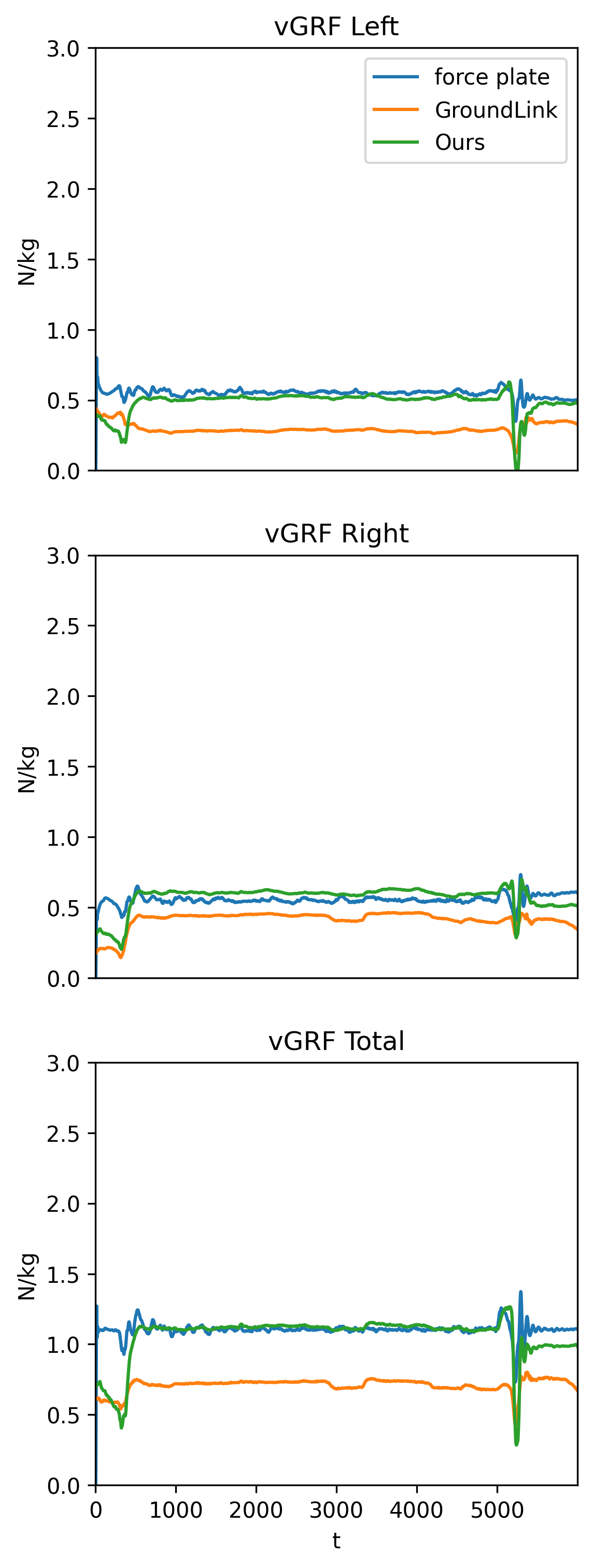}
        \caption{Chair}
        \label{subfig:chair}
    \end{subfigure}
    ~
    \begin{subfigure}[h]{0.23\textwidth}
        \centering
        \includegraphics[width=\linewidth]{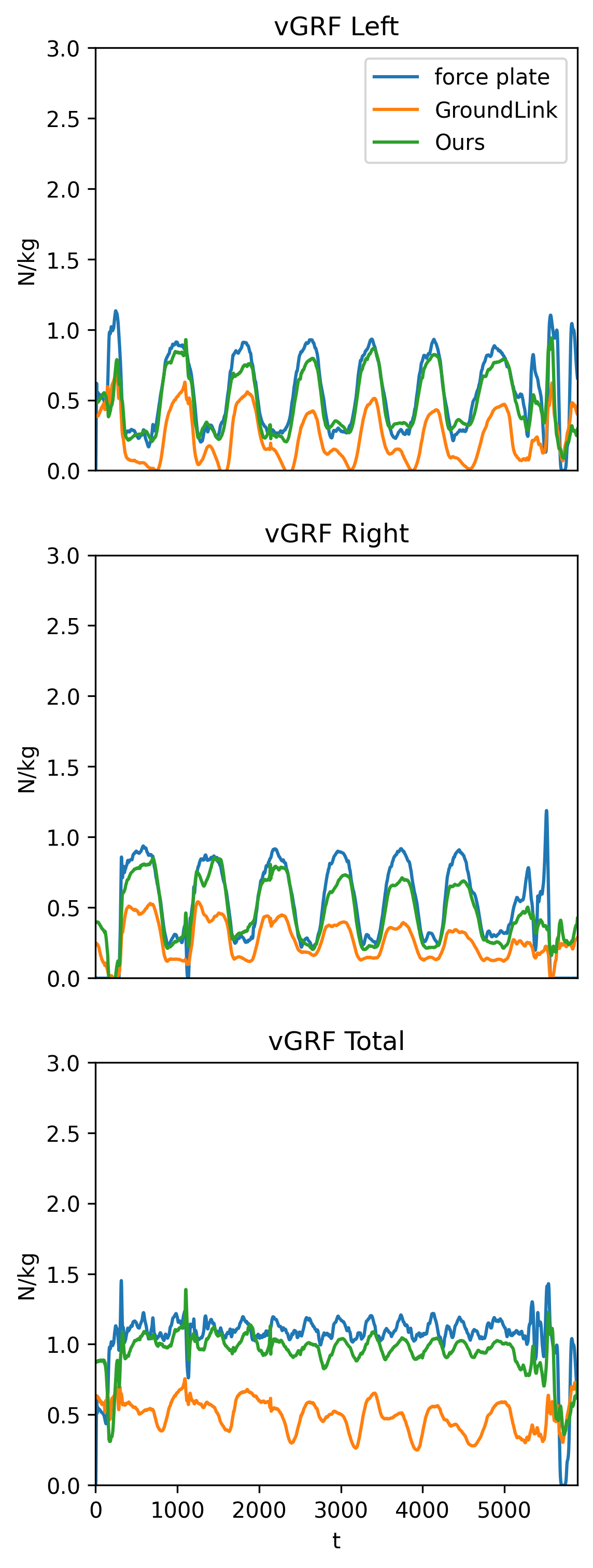}
        \caption{Side Stretching}
        \label{subfig:stretch}
    \end{subfigure}
    \caption{A few qualitative examples of the proposed model's predictions in comparison with the baseline \cite{han_groundlink23}. Our model's prediction of ground reaction forces (green) match the measurements from the force plates (blue) with higher accuracy that the GroundLink (orange). This is due to the additional of physics-informed learing objective in the training phase. Our pipeline provides higher robustness and long-term reliability in addition to learning approaches with only force plate data.}
    \label{fig:exp_graphs}
\end{figure*}

\subsection{Comparison to the baseline}
\label{subsec:results}

We report the quantitative evaluation results of the proposed approach in Tab. \ref{tab:vGRF_results} in comparison with the related works. As suggested by \cite{mourot_cgf22}, the vertical ground reaction force (vGRF) provides the most informative evaluation, we report our results on the mean squared error (MSE) of vGRF between the model's prediction with force plates data, in comparison with UnderPressure \cite{mourot_cgf22} and GroundLink baseline \cite{han_groundlink23}. As can be seen in Tab. \ref{tab:vGRF_results}, the proposed approach achieve state-of-the-art on the vGRF metric, especially on the vGRF estimation of left foot contact, with $18\%$ of improvement on average to the baseline GroundLink model. A few individual samples are shown in Fig. \ref{fig:exp_graphs}, qualitatively demonstrating the performance of our approach. Compared to the baseline \cite{han_groundlink23}, our model matches the force plates measurement with higher accuracy, due to contribution of the physics-based information during the training phase.

Additionally, we evaluate the vertical root position error (vRPE) results given the prediction of ground reaction forces from the \emph{GroundLinkNet} model. The simulation process follows Eq. \ref{eq:eom} and Eq. \ref{eq:euler}, replacing PD-controlled $\mathbf{F}$ with the model prediction $\hat{\mathbf{F}}$. The reported results are shown in Tab. \ref{tab:vRPE_results}, which are the MSE between the simulated trajectory and the input motion capture. Our model trained with physics-based loss in Eq. \ref{eq:loss} significantly outperforms the baseline GroundLink on most evaluation motions ($38\%$ error decrement on average). The 'walking' sample demonstrated in Fig. \ref{subfig:walk} results in a poor trajectory reconstruction with vRPE of $75.59$. With our approach, the sample has sufficient knowledge about the ground dynamics despite the missing of force plate data, resulting in a better trajectory with vRPE of $7.9$.

\subsection{Ablation studies}
\label{subsec:ablation}

We conduct two ablation studies to verify the proposed approach. The first study is the impact of having physics-based objective towards the learning of human reactive dynamics. To examine this quantity, we varies the loss weight $w_2$ in Eq. \ref{eq:loss} from $0.002$ to $0.010$ with step of $0.002$, while keeping the weight $w_1=0.002$ unchanged as provided by GroundLink \cite{han_groundlink23}. The results are illustrated in Tab. \ref{tab:ablation_weight}. As can be seen, the model trained with a weight of $0.005$ achieves the best evaluation results on vGRF of both left and right foot contacts (vGRF\_L and vGRF\_R) and vRPE, thus being the chosen learning parameter for the comparison in Tab. \ref{tab:vGRF_results} and Tab. \ref{tab:vRPE_results}. 

\begin{table}[t]
    \centering
    \caption{The ablation study on the impact of the learning weight $w_2$ on the prediction of \emph{GroundLinkNet} \cite{han_groundlink23}. Higher $w_2$ leads to a reduction in simulated root trajectory accuracy.}
    \begin{tabularx}{\linewidth}{@{\extracolsep{\fill}}l|ccccc}
    \toprule
    Model & $w_1$ & $w_2$   & vGRF\_L & vGRF\_R & vRPE \\
    \midrule
    \emph{GLinkNet} \cite{han_groundlink23} & $0.002$ & - & $0.11$ & $0.06$ & $38.43$ \\
    Ours & $0.002$ & $0.001$ & $0.09$ & $0.06$ & $19.58$ \\
    Ours & $0.002$ & $0.002$ & $0.09$ & $0.06$ & $16.23$ \\
    Ours & $0.002$ & $0.003$ & $0.10$ & $0.06$ & $16.15$ \\
    Ours & $0.002$ & $0.004$ & $0.10$ & $0.06$ & $16.23$ \\
    Ours & $0.002$ & $0.005$ & $0.09$ & $0.06$ & $\bold{14.69}$ \\
    \bottomrule
    \end{tabularx}
    \label{tab:ablation_weight}
\end{table}

The second ablation study is the parameter selection process of the PD controller in Eq. \ref{eq:pd}. The total ground reaction force is designed to be responsible for the root translation, therefore its magnitude is proportional to the offset between root position from two consecutive time steps. We conduct the experiment across all participants of the GroundLink dataset and the selected parameters are based on the average vRPE score. The proportional control parameter $\kappa_P$ is varied from $10$ to $90$ with a step of $20$. As can be seen in \ref{tab:ablation_pd}, the best $\kappa_P$ is recorded at $70$. However, with only the proportional term, a smooth trajectory cannot be easily achieved. We varies the dampening gain $\kappa_D$ from $3$ to $15$ with a step size of $3$, and the reconstruction accuracy significantly reduce. The best average vRPE score of $0.32 \pm 0.06$ is recorded at $\kappa_P = 70, \kappa_D = 3$, and they are implemented as the chosen parameters for the evaluation results in Tab. \ref{tab:vGRF_results} and Tab. \ref{tab:vRPE_results}.

\begin{table}[t]
    \centering
    \caption{The ablation study on the PD control parameters. The results are the MSE in meters between PD-simulation and original root position, scaled by a factor of $10^3$. We underscore the best simulation result when using only the proportional term. With the dampening support from the derivative term, the best simulated trajectory is recorded at $\kappa_P = 70, \kappa_D = 3$.}
    \scriptsize
    \addtolength{\tabcolsep}{-0.8em}
    \begin{tabularx}{\linewidth}{@{\extracolsep{\fill}}cc|cccccccc}
    \toprule
    $\kappa_P$ & $\kappa_D$ & S1 & S2 & S3 & S4 & S5 & S6 & S7 & Avg \\
    \midrule
    $10$ & $0$  & $88.67$ & $77.01$  & $235.68$ & $118.11$ & $54.29$  & $57.02$  & $89.72$  & $102.92 \scriptstyle{\pm 62.41}$ \\
    $30$ & $0$  & $49.30$ & $264.91$ & $157.13$ & $16.03$  & $106.91$ & $123.94$ & $386.22$ & $157.77 \scriptstyle{\pm 128.61}$\\
    $50$ & $0$  & $3.24$  & $5.71$   & $3.84$   & $3.11$   & $1.62$   & $2.26$   & $4.26$   & $3.43 \scriptstyle{\pm 1.34}$ \\
    $70$ & $0$  & $1.22$  & $4.86$   & $2.28$   & $1.84$   & $3.01$   & $2.26$   & $4.03$   & $\underline{2.78 \scriptstyle{\pm 1.27}}$ \\
    $90$ & $0$  & $0.70$  & $12.21$  & $5.11$   & $1.03$   & $0.60$   & $1.87$   & $9.02$   & $4.36 \scriptstyle{\pm 4.63}$ \\
    $70$ & $3$  & $0.26$  & $0.42$   & $0.35$   & $0.27$   & $0.29$   & $0.33$   & $0.37$   & $\bold{0.32 \scriptstyle{\pm 0.06}}$ \\
    $70$ & $6$  & $0.28$  & $0.45$   & $0.39$   & $0.30$   & $0.34$   & $0.37$   & $0.39$   & $0.36 \scriptstyle{\pm 0.06}$ \\
    $70$ & $9$  & $0.34$  & $0.56$   & $0.48$   & $0.37$   & $0.43$   & $0.44$   & $0.48$   & $0.44 \scriptstyle{\pm 0.07}$ \\
    $70$ & $12$ & $0.40$  & $0.69$   & $0.58$   & $0.46$   & $0.53$   & $0.52$   & $0.59$   & $0.54 \scriptstyle{\pm 0.09}$ \\
    $70$ & $15$ & $0.47$  & $0.82$   & $0.68$   & $0.56$   & $0.62$   & $0.61$   & $0.70$   & $0.64 \scriptstyle{\pm 0.11}$ \\
    \bottomrule
    \end{tabularx}
    \label{tab:ablation_pd}
\end{table}

\section{Conclusion}
\label{sec:conclusion}

In this paper, we propose a physics-informed approach for the learning of ground reaction dynamics from human motions. Prior works often rely on the supervision from the measurements of the restricted laboratory sensors such as force plates as the mean of supervision for the learning task. Despite the valuable measurements, force plates data often contains noises and missing of data, especially when the actions are performed by inexperienced actors. To this end, we propose a novel approach to generate additional ground truth reaction dynamics data fully depends on the more reliable motion capture data instead. On the GroundLink dataset, the approach demonstrated a significant gains in prediction accuracy of ground reaction forces, along with a high performance on a plausibility metric such as trajectory simulation.

\textit{Discussion and Future Works}. Despite the accurate measurement of ground dynamics, the proposed approach only consider the root translation at the moment. Extending the modeling to root and body joint rotation requires more in-depth understanding about the human body, such as how the reaction moments propagate backward to every body joint, and this will be investigated in the future as the natural progression of the project. Moreover, the dataset is constrained to only two feet contacts, leading to a limited type of motions that can be studied. The modeling of more complex contacts with the ground and the surrounding environment is an on-going and worth-wide line of study for computer vision, graphics and biomechanics research.

\noindent \textbf{Acknowledgments.} This research is supported by the Wallenberg Artificial Intelligence, Autonomous Systems and Software Program, from Knut and Alice Wallenberg Foundation.

\bibliographystyle{ieeetr}
\bibliography{references}
\end{document}